\documentclass[final,5p,times,twocolumn]{elsarticle}

\usepackage[ruled,linesnumbered]{algorithm2e}
\usepackage{graphicx}
\usepackage{subfigure}
\usepackage{arydshln}
\usepackage{amssymb}
\usepackage{tabularx}
\usepackage{booktabs}
\usepackage[pdftex,table,fixpdftex]{xcolor}
\usepackage{multirow}

\newcommand{\MM}[1]{\ensuremath{\mathbf{M}{#1}}}
\newcommand{\MT}[1]{\ensuremath{\mathbf{T}{#1}}}
\newcommand{\MW}[1]{\ensuremath{\mathbf{W}{#1}}}
\newcommand{\MWback}[1]{\ensuremath{\mathbf{W^{back}}{#1}}}
\newcommand{\MWin}[1]{\ensuremath{\mathbf{W^{in}}{#1}}}
\newcommand{\MWjin}[1]{\ensuremath{\mathbf{W^{in}_\mathnormal{j}}{#1}}}
\newcommand{\MWj}[1]{\ensuremath{\mathbf{W_\mathnormal{j}}{#1}}}

\newcommand{\MWout}[1]{\ensuremath{\mathbf{W^{out}}{#1}}}
\newcommand{\MWjout}[1]{\ensuremath{\mathbf{W^{out}_\mathnormal{j}}{#1}}}
\newcommand{\Md}[1]{\ensuremath{\mathbf{d}{#1}}}
\newcommand{\Mf}[1]{\ensuremath{\mathbf{f}{#1}}}
\newcommand{\Mu}[1]{\ensuremath{\mathbf{u}{#1}}}
\newcommand{\Mx}[1]{\ensuremath{\mathbf{x}{#1}}}
\newcommand{\My}[1]{\ensuremath{\mathbf{y}{#1}}}
\def \ti {\it t}

\definecolor{cornsilk}{rgb}{1,0.97,0.86}
\definecolor{cornsilk2}{rgb}{0.93,0.91,0.83}

\journal{not yet}

\begin{document}

\newcolumntype{L}{>{\columncolor{yellow!20}}l}
\newcolumntype{R}{>{\columncolor{yellow!20}}r}

\begin{frontmatter}

\title{Distributed Fault Detection in Sensor Networks using a Recurrent Neural Network}

\author{Oliver Obst}
\ead{oliver.obst@csiro.au}
\address{CSIRO ICT Centre, Autonomous Systems Laboratory \\
Locked Bag 17, North Ryde, NSW 1670 - Australia}


\begin{abstract}
In long-term deployments of sensor networks, monitoring the quality of gathered data is a critical issue.
Over the time of deployment, sensors are exposed to harsh conditions, causing some of them
to fail or to deliver less accurate data. If such a degradation remains undetected, the usefulness of
a sensor network can be greatly reduced. We present an approach that learns spatio-temporal 
correlations between different sensors, and makes use of the learned model to detect misbehaving
sensors by using distributed computation and only local communication between nodes.
We introduce SODESN, a distributed recurrent neural network architecture, and a learning method to train
SODESN for fault detection in a distributed scenario. Our approach is evaluated using data from
different types of sensors and is able to work well even with less-than-perfect link qualities 
and more than 50\% of failed nodes.

\end{abstract}


\end{frontmatter}


\section{Introduction}

Wireless sensor networks (WSN) are increasingly being deployed over extended periods of 
time \cite{CVS+07}, in particular for environmental monitoring applications. To facilitate long term deployments in remote areas, nodes are typically powered by solar energy and rechargeable batteries. 
Consequently, much of the research has focussed on energy-aware design of hard- and software as 
well as on building models of energy supply and demand. The continuing progress in this area has 
lead to longer autonomy of WSN,
but also revealed that deploying a sensor network over a long period of time requires automatic
monitoring of the quality of gathered data and of the condition of solar panels, sensors and batteries.
With information about the performance of these components, maintenance trips to remote monitoring 
sites can be better planned or possibly avoided, leading to a reduction of management costs. 
Some of the faults might be easier to detect than others: when some of the expected data is missing,
fault seem obvious to recognize. Even in this simple case, an automatic notification relieves the 
administrator from continuously monitoring a database. When the network delivers data as expected, 
there might also be more subtle problems, like mis-calibration or build-up of dust on sensors and solar 
panels, leading to incorrect sensor readings or shorter duty-cycles and thus less data. 
To prevent this, sensor networks have to 
become more user-friendly: existing systems often require to manually detect and diagnose potential
problems. First steps towards higher reliability and user-friendliness are automatically building a model
of the normal system behavior and to use this model to detect anomalies. With the result of this
process, it is possible to notify administrators who then can decide on appropriate actions. Consequently, 
the system can run unobserved with less danger of losing important data. 

For this work, we are interested in detecting problems that manifest in changes of sensor readings for 
some of the nodes of an entire network as a result of a sensor fault. 
Typically, some of the sensors at different nodes are correlated over space or time. We present an 
approach that is able to learn spatio-temporal correlations and make use of
them for detecting anomalies in a decentralized way, without using global communication during normal
operation. Instead, sensor nodes participate in a large, distributed recurrent neural network, where 
each of the sensor nodes hosts only a few neural units and communicates only with its local neighbor 
sensor nodes. Our neural network approach is inspired by echo state networks (ESN)~\cite{JH04}, 
a recurrent neural network approach which 
has shown to be successful in learning even complex time series. 
ESN have already been applied in anomaly detection in sensor networks~\cite{OWP08}, but  
only in a way that requires one instance of an ESN on each node. 
This results in an unnecessary consumption of memory resources and processing power.
A straightforward distribution of an ESN over the entire sensor network is also not a solution, because it requires all of the nodes to communicate with each other. More often than not, this sort of communication is 
neither available nor desired in sensor networks. 

To address the problem of detecting sensor faults in WSN in a distributed way, we introduce spatially 
organized distributed echo state networks (SODESN), an architecture 
that allows for distributing a single recurrent neural network over an entire sensor network even when the 
WSN imposes a local communication structure on its connectivity matrix (Sect.~\ref{sec:desn}). In
Sect.~\ref{sec:learning}, we present a training method for SODESN and an approach to train SODESN
for fault detection in WSN. SODESN learn a model of normal behavior of sensor nodes based on 
information from other sensors. The fault detection in turn monitors differences between the model and 
actual sensor readings in a distributed way.
We demonstrate the capabilities of our approach with data from different 
temperature and radiation sensors (Sect.~\ref{sec:exp}) and discuss our results in 
Sect.~\ref{sec:discussion}. In the following section, we start with a brief overview of related work, and
give a short review of the ESN approach, the starting point for our work.

\section{Background}
\label{sec:background}

Detecting and diagnosing faults is a challenge that has been addressed in many different areas for 
different purposes. Logic-based approaches, for instance, can be applied if a complete description 
of the desired behavior of the system is available (see e.g.~\cite{Rei87}). 
In distributed systems, approaches like in~\cite{BSM89} detect faults by using connections between 
processors to implement a voting based diagnosis system. 
WSN are distributed systems where different components, from batteries over sensors to processors,
contribute to potentially many different types of faults. There may be problems with the energy supply, 
with the routing or other communication problems, resulting in missing data from single nodes, or 
causing the whole system to deliver no data at all. In long-term deployments, problems like degradation of 
hardware can result in inaccurate measurements, caused by dust and continued exposure of sensors to 
the environment. Some of the existing work tackles the problem of automatically detecting node failures
with centralized approaches (e.g.~\cite{RNL03}), where relevant information is forwarded to a dedicated 
manager performing the fault detection. 
Methods to detect faults in a distributed way have been investigated, because global communication
becomes prohibitive with increasing network sizes. The approach in~\cite{CKS06}  is an example of such
a decentralized approach, where sensor faults are detected based on differences in the readings between 
neighbors. It uses only local communication between nodes, but assumes that all sensors measure 
the same variable. Likewise, \cite{RNEV08} is able to detect faults with a distributed approach, but 
here, the assumptions are not as strong. Neighbor sensors are not required to measure the same 
variable, but are assumed to be correlated as long as they are working normally, and uncorrelated
as soon as they are faulty. This fault detection method uses a graph-based approach to isolate faulty 
nodes in the network, where correlation between the time series over a time window is used to identify 
faults.

In our work, we are also interested in detecting sensor faults in a distributed way. Instead of explicitly
basing our fault detection on spatial correlations between sensors, we want our system to detect the 
relevant spatio-temporal correlations on its own. If we are able to distribute a
large recurrent neural network over the entire sensor network, each sensor node can estimate its own 
true values based on information from its neighbors in a training period. Because recurrent neural 
networks model dynamical systems (i.e.\ with a memory of past events), correlations can be both 
temporal as well as spatial. Using the estimated true values, and a threshold on deviation between 
estimated value and recent readings, each node can decide if it can be assumed to work correctly.

\subsection{ESN technical background}
\begin{figure}[htbp]
\centering
\includegraphics[width=0.45\textwidth]{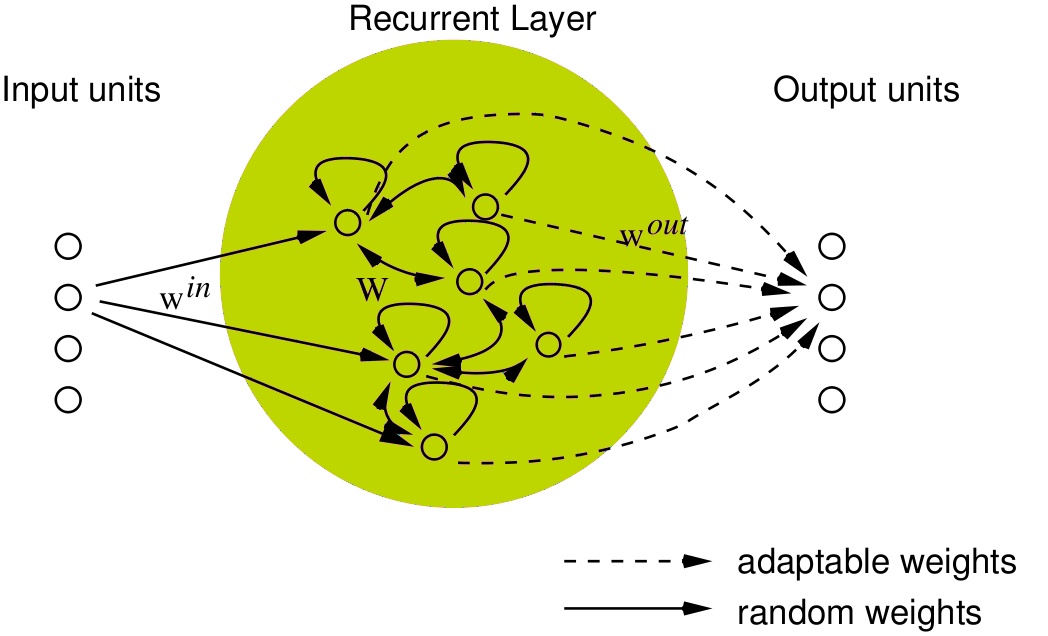}
\caption{Echo State Network.}
\label{fig:esn}
\end{figure}

Recurrent neural networks have only recently become more widely used in practice, because many 
approaches have been difficult to set up and to train for. An ESN is a specific type of recurrent neural 
network which is able to successfully predict complex time series~\cite{JH04}. At the same 
time, the complexity of training an ESN is much lower than with traditional recurrent neural networks. 
Like any other neural network, ESN consist of neural units and synaptic connections between these units. A 
neural network is recurrent if there is at least one cycle in these connections. Units are typically organized in 
different layers and possess a state (called ``activation''). This activation is computed (using a typically 
non-linear  ``activation function'') based on inputs from incoming connections. Connections between units perform a linear transformation and can be either excitatory (positive connection weights) or inhibitory 
(in case of negative connection weights).
Traditional approaches to training recurrent neural networks, like backpropagation through time~\cite{Wer90}, 
change all of the weights between different units. 
The lower training complexity of ESN is a result of using a fixed, randomly connected ``reservoir'' of neural
units in the recurrent layer,  and only changing connections to output units during training (see Fig.~\ref{fig:esn}). 
Once the training is finished, connections are changed no longer. Both output and the next state of the 
network are determined by the current state of the network and the current input.

To make the approach work, however, 
connections cannot be entirely random, but need to fulfill the so-called \emph{echo state condition} \cite{Jae02}. 
For an illustration of this condition \cite{MB04}, consider a time-discrete recursive function ${\Mf{} x}_{\ti +1} = F({\Mf{} x}_{\ti  },  {\Mf{} u}_{\ti } )$ that is defined at least on a compact sub-area of the vector-space ${\Mf x} \in R^n$, with $n$ the number of internal units. The 
${\Mf x}_{\ti }$ are to be interpreted as internal states and ${\Mf u}_{\ti }$ is some external input sequence. 
Now, assume an
infinite input sequence: $\bar{\Mf u}^{\infty} = {\Mf u}_0, {\Mf
  u}_1, \dots$ and two random initial internal states of the system
${\Mf{} x}_0$ and ${\Mf{} y}_0$.  To both initial states ${\Mf{} x}_0$ and
${\Mf{} y}_0$ the sequences $\bar{\Mf{} x}^{\infty} = {\Mf{} x}_0, {\Mf{}
  x}_1, \dots$ and $ \bar{\Mf{} y}^{\infty} = {\Mf{} y}_0, {\Mf{} y}_1,
\dots$ can be assigned.
\begin{eqnarray}
{\Mf{} x}_{\ti+1} = F({\Mf{} x}_{\ti },{\Mf{} u}_{\ti } ) \label{eqn:sesn1}\\
{\Mf{} y}_{\ti+1} = F({\Mf{} y}_{\ti },{\Mf{} u}_{\ti } ) \label{eqn:sesn2}
\end{eqnarray}

The system $F(\cdot)$ fulfills the echo state condition if it is independent from the set ${\Mf{} u}_{\ti }$, and if for any (${\Mf{} x}_{0}$,${\Mf{} y}_{0}$) and all real values $\epsilon > 0$, there exists a $\delta(\epsilon)$ for which
$d({\Mf{} x}_{\ti }, {\Mf{} y}_{\ti }) \leq \epsilon$ for all $ \ti  \geq \delta(\epsilon)$, where $d$ is a square Euclidean 
metric. Two rules are a helpful for creating a connectivity matrix $\MW{}$ with this condition:
\begin{description}
\item[C1] it is a necessary condition that the spectral radius of the biggest 
eigenvalue of \MW{} is below one.
\item[C2] it is a sufficient condition that the biggest singular value of \MW{} is smaller than one
\end{description}

Using one ESN for each sensor node, or one ESN in a central location, would require a combination of 
high memory resources on each node, an explicit selection of correlated sensors or global communication.
Instead, we describe a new approach where we distribute a recurrent neural network over an entire 
sensor network, fulfill the above mentioned echo state condition, and use only communication 
between neighbor nodes. 

With sensor networks and recurrent neural networks, two different kinds of networks play a role in the following. In order to avoid confusion between the two in our description, we use \emph{node} when 
we talk of sensor network nodes, whereas we use \emph{unit} for the components of a neural network. In our 
notation we use bold capital letters for matrices, bold small letters for vectors or vector-sized functions, and italics for  scalars.

\section{Spatially organized distributed echo state networks}
\label{sec:desn}

To distribute a recurrent neural network over a WSN, connections between units have to be 
restricted to the spatial neighborhood of sensor nodes in order to avoid unrestricted global communication.
We also would like to retain the efficient learning of ESN. Therefore, we create neural units on each sensor 
node, and follow the original idea of ESN in that all connections between internal units are randomly initialized 
and fixed. Connecting units only to spatial neighbors on different devices leads to our idea of
spatially organized distributed echo state networks (SODESN), where the underlying communication structure 
of the sensor network prohibits to use arbitrary synaptic connections between distributed units.
More specifically, we allow hidden units to be connected to each other only if they are hosted on the same 
or on a neighbor network node. Moreover, neural inputs are only connected to units on the same sensor node 
in order to further reduce communication. Instead of globally connected output units, we use local output units 
on each sensor node. Output units get their input from the local part of the reservoir and from reservoirs on 
neighbor nodes. 

ESN typically use a sparsely connected reservoir, so that different internal units develop different dynamics.
Outputs are then calculated as a linear combination of the (non-linear) internal units. Using only local 
connections in SODESN almost automatically leads to a sparse connection matrix, albeit with 
a different distribution of connections. From a global perspective, regarding a SODESN as a single neural
network, we also want to make sure the system fulfills the echo state condition mentioned in the 
previous section.

In a setup with $M$ sensor nodes, each node $m$ hosts $K_m$ input units, $N_m$ hidden units, and $L_m$ output units. The total number of neural units thus is
\begin{eqnarray*}
  K & = & \sum_{m=1}^M K_m \mathrm{~inputs,~}   N = \sum_{m=1}^M N_m \mathrm{~hidden~units,~and~}  \nonumber \\
  L & = & \sum_{m=1}^M L_m \mathrm{~output~units.} 
\end{eqnarray*}

Then, from a global perspective, the SODESN model consists of $K$ input units with an activation 
vector 
\begin{equation}
\mathbf{u}(n) = (\underbrace{u_{1_1}(n),...,u_{K_1}(n)}_{\mathrm{node~1}}, 
           ~...,~ 
           \underbrace{u_{1_M}(n),...,u_{K_M}(n)}_{\mathrm{node~}M})',
\end{equation}

of $N$ hidden units with an activation vector
\begin{equation}
\mathbf{x}(n) = (x_{1_1}(n),...,x_{N_1}(n), 
           ~...,~ 
           x_{1_M}(n),...,x_{N_M}(n))',
\end{equation}

and of $L$ output units with an activation vector
\begin{equation}
\mathbf{y}(n) = (y_{1_1}(n),...,y_{L_1}(n), 
           ~...,~ 
           y_{1_M}(n),...,y_{L_M}(n))'.
\end{equation}

For the rest of this paper, we assume all neural units to be evenly distributed over all sensor nodes, i.e.\ 
each node contains the same number of units.
 
For theoretical considerations, it is convenient to represent synaptic connections weights between 
units in several global matrices, which have to be distributed in a practical implementation. 
Connections between hidden units 
are represented in a $N \times N$ matrix $\mathbf{W} = (w_{ij})$, connections from
input units to hidden units in a $N  \times K$ matrix 
$\mathbf{w^{in}} = (w_{ij}^{in})$ , and connections from input and hidden units to output units in a 
$L \times (K+N)$ matrix $\mathbf{w^{out}} = (w_{ij}^{out})$.

The activation of internal units is computed as 
\begin{equation}
\Mx{(n+1)} = \Mf{(\MWin{} \Mu{(n+1)} + \MW{} \Mx{(n)})},
\end{equation}

where $\mathbf{u}(n+1)$ represents the readings from all sensors, and $\mathbf{f}$ the vector
of activation functions $f$ of all internal units. We use $f = \tanh$ as activation function in each internal unit,
and linear input- and output units ($f = 1$).  In some cases, ESN use connections projecting back from outputs 
into the reservoir. This is also possible in SODESN and requires an additional matrix \MWback{}.
Consequently, the activation of internal units \Mx{(n+1)} is then  computed as 
$\Mf{(\MWin{} \Mu{(n+1)} + \MW{} \Mx{(n)} + \MWback{} \My{(n)})}$. For our application, we do not make use 
of these connections.


\begin{figure*}[htbp]
\centering
\includegraphics[width=\textwidth]{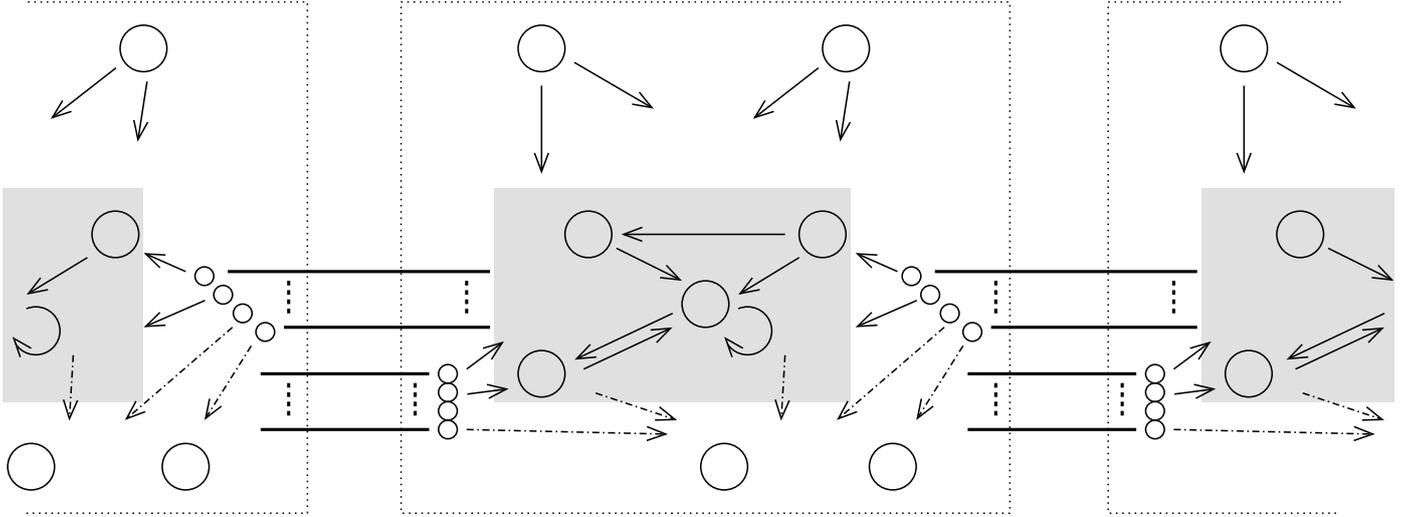}
\caption{Neural units in a sensor node and connections to units on neighbor nodes.}
\label{fig:structure}
\end{figure*}

\subsection{Proxy units}
In a practical implementation, activation vectors are distributed over multiple sensor nodes.
Moreover, there are connections between units on different sensor nodes, which require to have a
specified physical location. We store incoming connections from units hosted on neighbor sensor nodes 
on the local node. Units with outgoing connections to units on other devices just
forward their activations with no changes to the neighbor device. Additional proxy units on the neighbor
act as a place holder for remote units and take activations from connected units. 
From proxy units, there are only local connections to the reservoir or to output units.
Proxy units also eliminate the need for all sensor nodes being synchronized as long as they all use the same
interval to process data (e.g.\ once every minute or every 15 minutes). After new activations have been 
computed, their values are forwarded to connected proxy units where they can be used by the neighbor 
device.  Once their values have been used, proxy units are reset to 0. This is to avoid using old values in 
case of a link failure between two network nodes. In our experiments described in Sect.~\ref{sec:exp}, 
we used link qualities of from $10\%$ to $100\%$.

\subsection{Initializing an untrained SODESN}

To set up the untrained SODESN, we construct the desired number of units on each sensor node. We 
create local internal connection matrices $\MW{}_j$ with a specified density, and scale each of them
so that the spectral radius is smaller than one. In addition, we create sparse random connections 
between internal units on neighbor devices, represented by connection from proxy units for incoming 
connections, and references to sensor nodes and respective proxy units for outgoing connections.
Local input connection matrices \MWjin{} with random weights fully connect input units to all local 
internal units on the node (with one input unit for each local sensor). For output units, we create local
random matrices \MWjout{} to provide them with input from input units, proxy units and internal units. 

The local internal connection matrices are scaled by their largest eigenvalue so that each spectral radius is 
at most one. For the entire connection matrix composed of all local matrices, this procedure does not in 
general lead to a spectral radius of smaller than one yet, but it leads to similar conditions for the 
internal units on each sensor node.
After all local matrices are created in this way, the resulting global connection matrix is scaled to meet
the echo state condition.

\begin{figure}[htbp]
\centering
\includegraphics[width=0.235\textwidth]{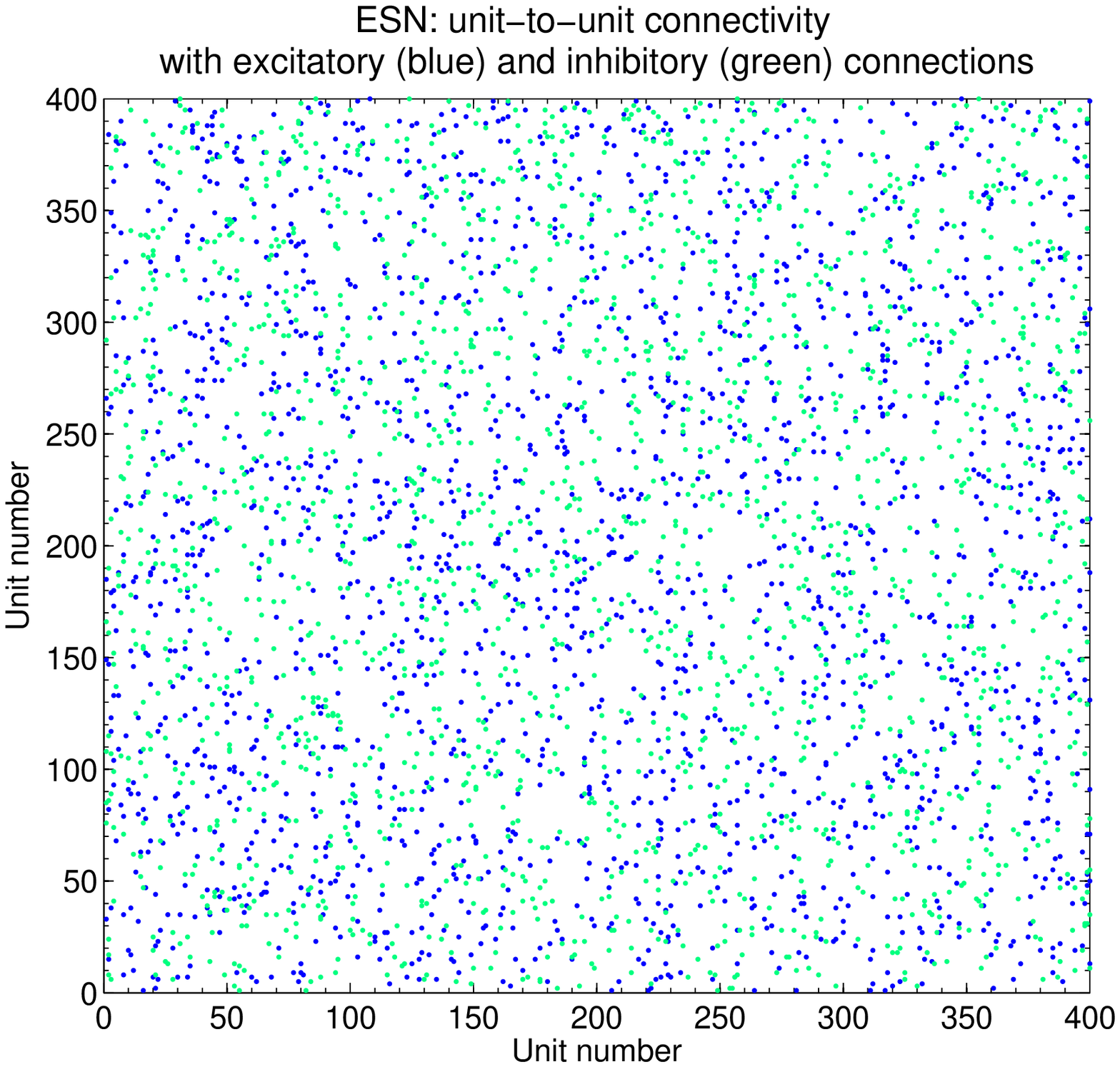} 
\includegraphics[width=0.235\textwidth]{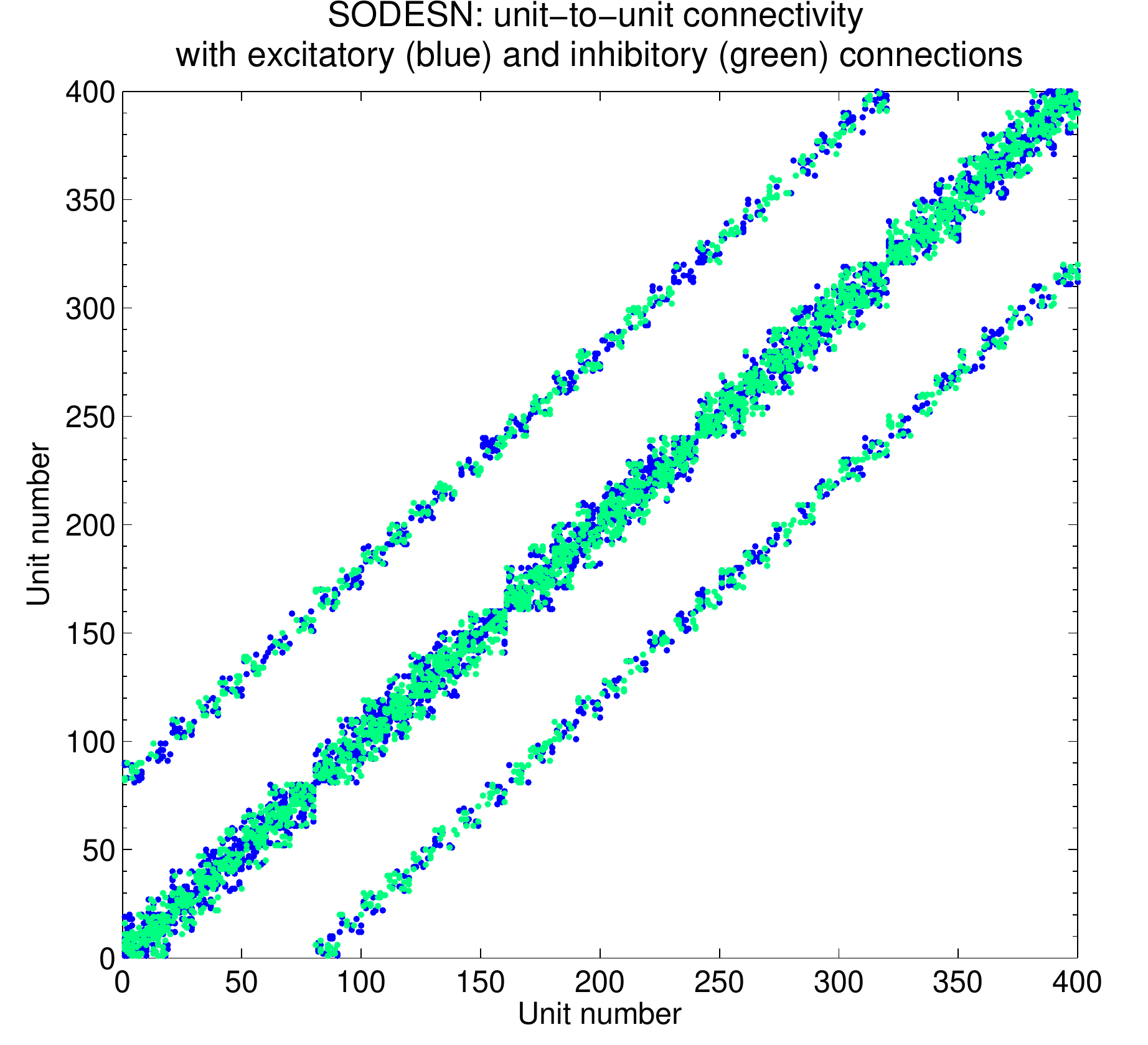}
\caption{Reservoir connectivity of a standard ESN (left) with 400 internal units and a SODESN (right), 
with 400 internal units distributed over a 5 $\mathbf{\times}$ 8 grid of sensor nodes.}
\label{fig:reservoirs}
\end{figure}

Algorithm~\ref{alg:setup} generates a distributed SODESN, where each sensor node hosts some input units, 
hidden units and output units. Globally, the sensor network imposes a 
specific structure on the random reservoir connectivity matrix. Figure~\ref{fig:reservoirs}
illustrates the difference in connectivity between a standard ESN and a SODESN.

\begin{algorithm}[htb] 
\caption{Initialization: on each node $j$ ...} 
\SetLine 
\dontprintsemicolon
Generate $K_j$ input units, $N_j$ internal units, and $L_j$ output units \;
Generate $M_j = \sum_i N_i$ proxy units for all neighbor sensor nodes $i$ as place holders for 
the internal units on neighbor nodes \;
For each neighbor sensor node $i$, create $N_j$ pointers to proxy units on node $i$ \;
Generate a sparse, random matrix  \MWj{} for connections between local internal units \;
Find  $\lambda_j$  as the largest eigenvalue of  \MWj{}  \;
Scale  \MWj{} by $1/\max(\lambda_j,1)$ \;
Choose $x \in \{0,...,1\}$, a connection density between neighbor units \;
Generate random connections from $x \times M_j{}$  of the local proxy units to local internal units \;
Generate and initialize an all zero $L_j \times (M_j + N_j + K_j)$ matrix for connections to local output units
from all other local units.
\label{alg:setup}
\end{algorithm}

\section{A training algorithm for SODESN}
\label{sec:learning}

After initial setup, SODESN needs to be trained. 
We describe an approach to offline training a SODESN in a supervised fashion, i.e.\ we need time series 
of both input and output units as training data. Once the training is finished, no further adaptation is made.
For our application to diagnose problems in sensor readings, we train output units to predict readings of a 
sensor in a neighbor node. In this case, the training data can be derived from any input time series of 
``normal'' sensor readings. 

\subsection{Offline training SODESN}

For a first description of the training algorithm, we regard SODESN as one recurrent neural network with specific 
connectivity -- from there, a distribution of the algorithm over all sensor nodes is straightforward. Unfortunately, 
the standard training approach for ESN (see~\cite{Jae02} for a detailed description) cannot be applied, because it 
assumes that output units can be connected to any of the input or the hidden units. In SODESN, we want to connect 
output units only to local input, internal or proxy units.

Training is executed in two steps, in a similar way to training ESN: as a first step, we sample a matrix \MM{} 
of internal network states, and a matrix \MT{} of output activations. Samples are taken while feeding a training data 
time series into input units (when using connections projecting back from output units into the reservoir, a 
teacher time series has also to be fed to output units). For each time step of the training data, 
we collect a vector of internal activations and a vector of output activations from our SODESN.
The sampled vectors are stored in new rows of \MM{} and \MT{}. 
With $N$ the total number of hidden units, $L$ the number of output units, and $S$  the number of training steps, 
the final sizes of \MM{} and \MT{} are $S \times N$ and $S \times L$, respectively. 
The first samples of a training are typically discarded in order to wash out the initial network state.

As a second step, we compute the output weights $w^{out}_{ij}$ to let the training time series 
$\mathbf{d}(n)$ for each output unit $j$ approximate a linear combination of the internal activations \Mx{(n)}.
``Approximate'' means to minimize the mean squared error on the training signal, which, in the case of
ESN, can be achieved by multiplying the pseudoinverse of \MM{} with \MT{}:
$(\MWout{})^t = \MM{^{-1}}  \MT{}.$
In SODESN, however, this operation is not possible, because it will create connections from all internal
units to all the output units. A solution to the problem is to adapt the output weights locally, by using 
local connection matrices \MM{_j} and \MT{_j} for each sensor node $j$. \MM{_j} contains only activations of local 
input, internal and proxy units, while \MT{_j} contains output activations of the local output units 
(see Algorithm~\ref{alg:training}). For each node, 
we compute a local output connection matrix:
\begin{equation}
(\MWjout{})^t = \MM{^{-1}_j}  \MT{}.
\end{equation}

An additional advantage of this operation, at least in theory, is that it can be performed on each sensor node in 
parallel. In many practical cases, however, the amount of desired training data and the complexity of the 
operation will exceed the available memory and limited processing power
of small sensor nodes. This is not a severe restriction, though, because the training needs to be done only once
and can be executed on a remote machine. The result of the training, a set of output weights, has then to be 
sent back to all nodes and installed in the local connection matrices.

\begin{algorithm}[tbp] 
\SetLine 
\dontprintsemicolon
\KwIn{$\mathbf{u}(n)$, $\mathbf{d}(n)$, $n = 0 ... T$, $T_0 < T$}

Initialize the network state $\mathbf{x}(0) = 0$ \;
\tcp{\small Sample network state for training series} \;
Initialize $\MM{} = \emptyset$, $\MT{} = \emptyset$ \;
\For {$n = 0...T$}{
$\Mx{(n+1)}$ =  \Mf{(\MWin{} \Mu{(n+1)} + \MW{} \Mx{(n)} )} \; 
\tcp{\small Discard initial states} \;
\If {$n >= T_0$}{ %
Add \Mx{} as a new row to \MM{} \;
Add $\tanh^{-1} \Md{(n)}$ as a new row to \MT{} \; %
}}
\tcp{\small compute sample matrices for each node} \;
\ForEach {\textup{sensor node} j}{ 
Initialize $\MM{}_j = \emptyset$, $\MT{}_j = \emptyset$ \;
\ForEach {\textup{column $\Mx{}'$ in \MM{}}}{
\If {$\Mx{}'$ are the activations of an internal unit on the same or on a neighbor node}{
Add $\Mx{}'$ as a new column to $\MM{}_j$ \; 
}
}
\ForEach {\textup{column $\My{}'$ in \MT{}}}{
\If {\textup{$\My{}'$ are the activations of an output unit on the same or on a neighbor node}}{
Add $\My{}'$ as a new column to $\MT{}_j$ \; 
}
}
\tcp{\small Compute all output weights for node j}\;
\tcp{\small using the pseudoinverse of $\MM{}_j$} \;
$(\MW{}^{out}_j)^t = \MM{}_j^{-1} \MT{}_j$ \;
}
\caption{Offline training SODESN} 
\label{alg:training}
\end{algorithm}

\newsavebox{\tempbox}%
\begin{figure*}[htp]
\centering
\sbox{\tempbox}{\includegraphics[width=0.5\textwidth]{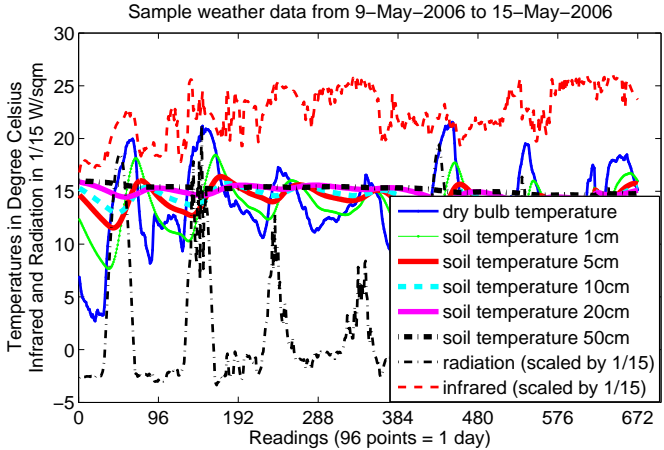}}
\begin{minipage}{0.50\textwidth}%
\subfigure[One week of the sensor data used in our experiments. In this graph, time series of infrared and radiation have been scaled by {$1\over{15}$}.]{\usebox{\tempbox}\label{fig:exampledata}}
\end{minipage}%
\hfill%
\begin{minipage}{0.47\textwidth}%
\subfigure[Arrows between sensor nodes indicate the local SODESN information exchange for fault detection.]{%
\vbox to \ht\tempbox{%
\vfil
\includegraphics[width=\textwidth]{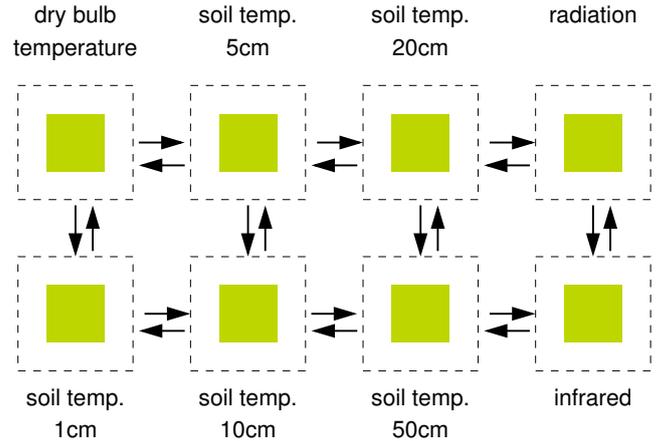}\label{fig:simsetup}%
\vfil }
}
\end{minipage}%
\caption{Training data and setup of the simulated sensor network. }
\label{fig:data}
\end{figure*}

\subsection{Training SODESN to detect sensor faults}

With the supervised training approach described above, we need to provide input as well as output signals 
for each sensor node. In our application to detect sensor faults, we expect the input signal and output signal 
for a sensor to be the same when the sensor works normally. To gather training data, the sensor network has 
to be deployed and collect sensor readings for a period of time. During this period, we assume there are no sensor 
faults, so that the training output for each sensor is exactly the same as the input time series. 

Using only normal data for training results in the learning to pick up this correlation. The output weights will be 
adjusted so that input and output always match closely. When a sensor is faulty and delivers unexpected
values to its input unit, the respective output will be similar to the input rather than an estimate of the true value.
In such a case, we cannot distinguish between normal or faulty sensors, so that our prediction is useless.

To fix the approach so that the prediction of the true value of one sensor is independent on its actual value,
a possible solution is to not connect this sensor to the neural network during both training and exploitation. The 
prediction is then solely based on inputs from other sensors. This is, however, only possible if we are interested
in monitoring just very few sensors in the network. To monitor all of the sensors, this would require to disconnect 
all of the sensors from the neural network. With no remaining inputs, we cannot make any predictions, so that this 
approach is not an option.

A more promising attempt is therefore to make only the training of one output unit independent of the respective
input unit. This can be achieved by training one output unit at a time, and disconnecting the input unit we are 
trying to predict during the training. However, this approach leads to a further problem: the prediction will be 
based on the assumption that there is no input from the sensor in question. During normal operation the input 
signal of the sensor will be added to some of the internal units and lead to a change in the output. In our 
experiments we found the influence of the incoming signal large enough to make the prediction useless.

Instead of just disconnecting individual input units during training of their respective output units, we 
make sure there is an actual signal from all of the inputs. For the input of the sensor we are currently 
training, the signal should be uncorrelated to the true sensor value. This can be achieved by for example
replacing the input by a white noise signal. The correct signal is used as teacher output, and the goal of 
the training is to learn the correlation between the true local sensor value and the value of neighbor sensors.

As mentioned above, the training aims to minimize the mean square error on the training signal.
In all our experiments, we tested the capability of the SODESN to generalize for new data by 
computing the normalized root mean square error (NRMSE) of the predictions on an independent test set.
The NRMSE of $n$ predictions $p$ of the SODESN against the test data $t$ is defined by
\begin{equation}
\mathrm{NRMSE} = \sqrt{{\sum_{i=1}^n (t(i)-p(i))^2} \over {n \ \mathrm{var}(t)} },
\end{equation}
where $\mathrm{var}(t)$ is the variance of the test data.

\newsavebox{\tempboxii}%
\begin{figure*}[htp]
\centering
\sbox{\tempboxii}{\includegraphics[width=0.32\textwidth]{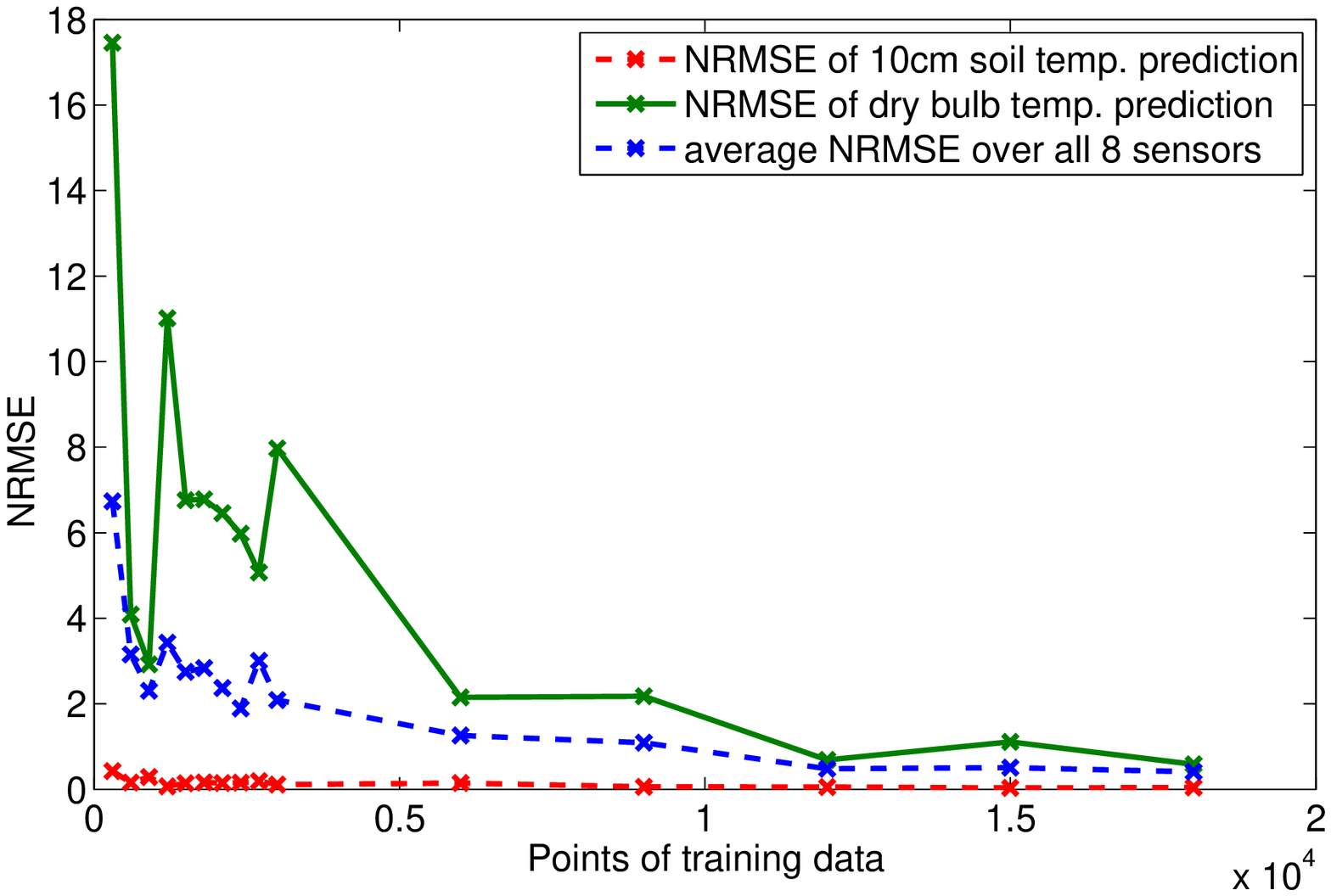}}
\begin{minipage}{0.32\textwidth}%
\subfigure[Learning curves for predicting the 10cm soil temperature, air temperature, 
and an average learning curve over all sensors. Experiments were run with a 90\% WSN connectivity
and training set sizes of up to 30.000 data points. The graph shows results of sets of up to 18.000 points.
]{\usebox{\tempboxii}\label{fig:results1}}
\end{minipage}%
\hfill%
\begin{minipage}{0.32\textwidth}%
\subfigure[Influence of the number of internal units per node on the learning performance. On average (blue line), 
an increasing number of internal units decreases the NRMSE only slightly. Experiments were run using a training
set size of 30.000 and 90\% success rate in communication between nodes. ]{%
\vbox to \ht\tempboxii{%
\vfil
\includegraphics[width=\textwidth]{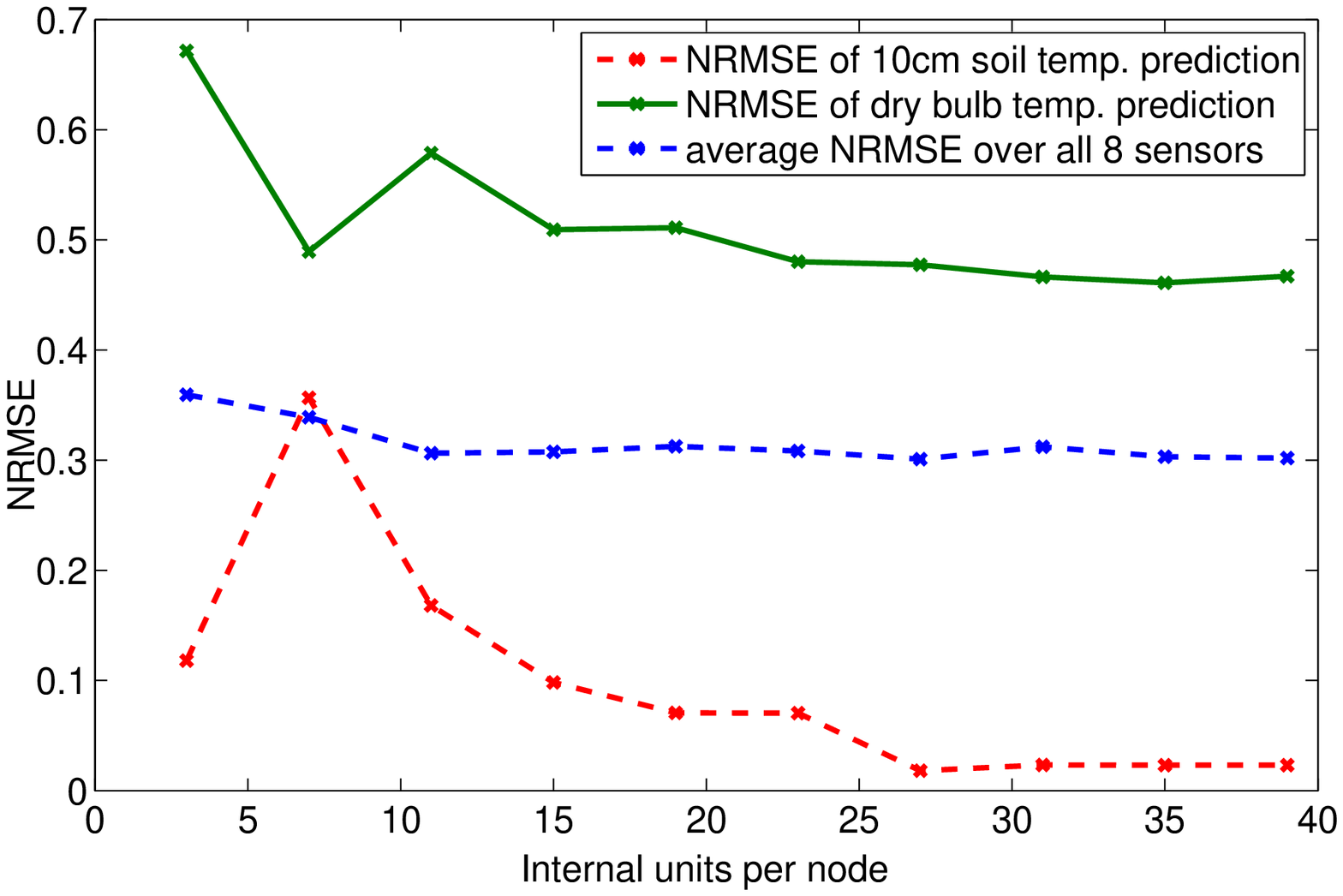}\label{fig:results2}%
\vfil }
}
\end{minipage}%
\hfill%
\begin{minipage}{0.32\textwidth}%
\subfigure[The benchmark against a centralized approach using one ESN for each prediction shows that
the SODESN is able to maintain a high prediction quality even with poor link qualities. Only under ideal 
conditions can the centralized approach keep up with SODESN. ]{%
\vbox to \ht\tempboxii{%
\vfil
\includegraphics[width=\textwidth]{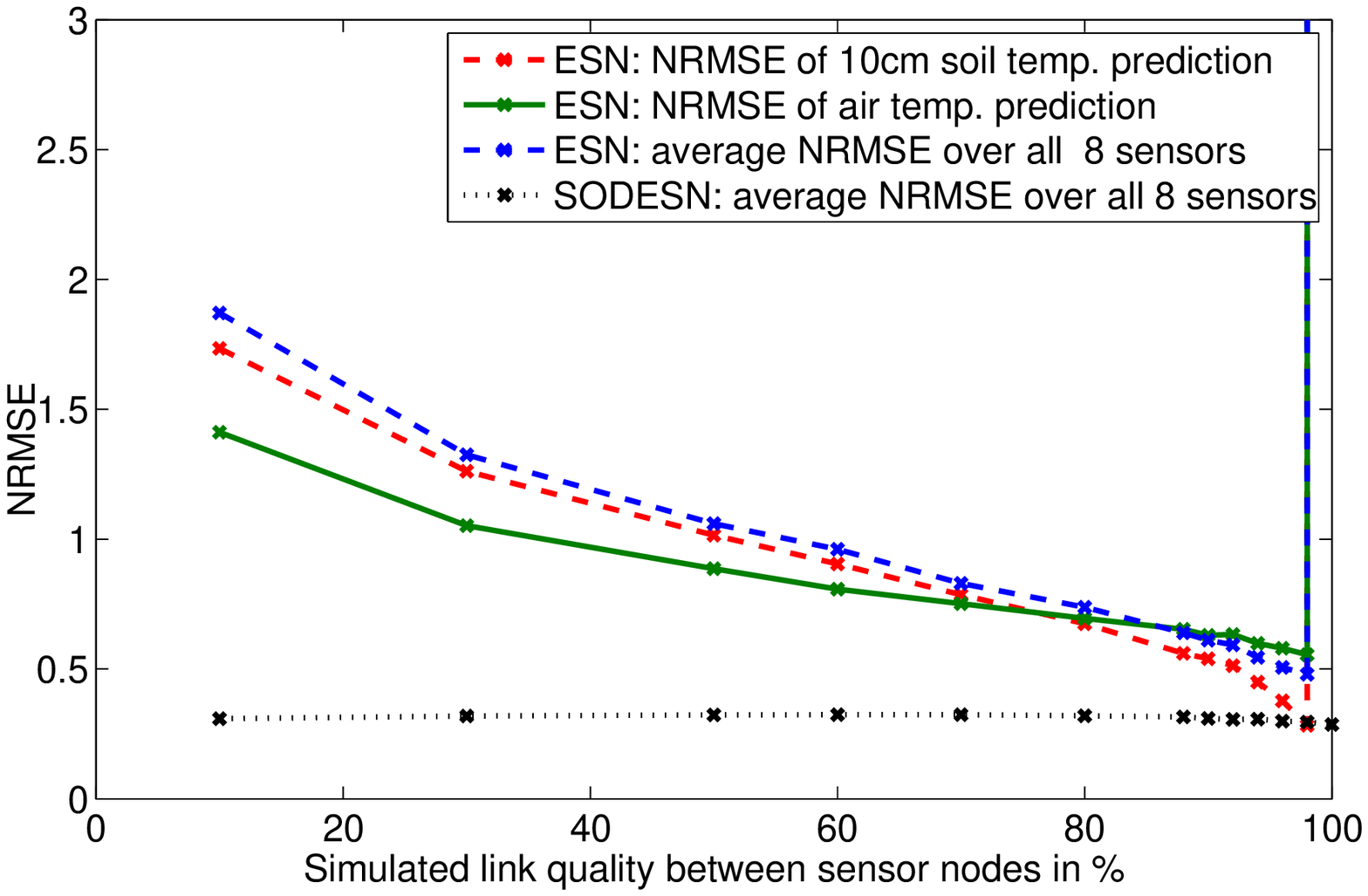}\label{fig:results3}%
\vfil }
}
\end{minipage}%
\caption{Results of various experiments using SODESN and a benchmark using ESN.}
\label{fig:results}
\end{figure*}

\subsection{Distributed fault detection}

Using SODESN for fault detection involves making predictions on each sensor node. 
It requires also to set a threshold for sensor readings to be considered 
abnormal. Possible methods for defining thresholds can be based on measuring deviations from the 
predicted value of a sensor (for example a deviation exceeding the maximum deviation of predictions 
on the test set), or on the NRMSE between prediction of the sensor value and its actual reading for a 
specified time window.

In the previous section, we set up the training so that predictions of a sensor are independent of its 
current value. By using random noise as local input during training, we base the fault  detection of each
sensor on input from the rest of the network.
If sensors fail only rarely, only a few of them will feed faulty values into the SODESN at the same
time. If there is a faulty sensor, it will continue to feed incorrect readings into
the network until the problem is fixed. This will affect fault detection in the remaining sensors,
even more so if more than one sensor is faulty at the same time.

In systems with a high likelihood of simultaneous sensor failures, it might therefore be a good idea 
to prevent faulty sensors from feeding their readings into the SODESN. For the same reasons we
used random noise as input during training in the previous section (as opposed to no input), we expect 
that simply disconnecting faulty sensors does not improve the predictions: after all,
output units in other nodes used a fraction of their input for training.
In order to decrease their effect on the system, we do flag and disconnect faulty sensors
from the SODESN. Instead of using no input from faulty sensors at all, we replace their input with
the predictions of their readings as computed by the SODESN. We expect this helping to maintain
a high prediction quality for the remaining sensors with a larger number of faults in the system.

\section{Experiments and results}
\label{sec:exp}

We evaluated our approach in simulations where we used data from a local weather station 
with several sensors measuring temperatures, radiation, infrared, etc. (the automatic weather station of
the Department of Physical Geography of Macquarie University~\cite{Mac08}).
The simulated setup consisted of 8 sensor nodes arranged
in a 2 by 4 grid where each node has one of the sensors and can communicate with its nearest neighbors 
(see Fig.~\ref{fig:simsetup}). The sensors we used measured the air temperature, soil temperatures 
at 1cm, 5cm, 10cm, 20cm, and 50cm respectively, radiation, and infrared. The data we used was taken in 
15 minute intervals.
Figure~\ref{fig:exampledata} shows data of our sensors for one week. In the graph it is visible that 
the different time series are at least weakly correlated to each other. In a setup with all sensors measuring the 
same variable at slightly different locations, correlations would be expected to be even stronger. 

\begin{figure*}[htb]
\centering
\includegraphics[width=\textwidth]{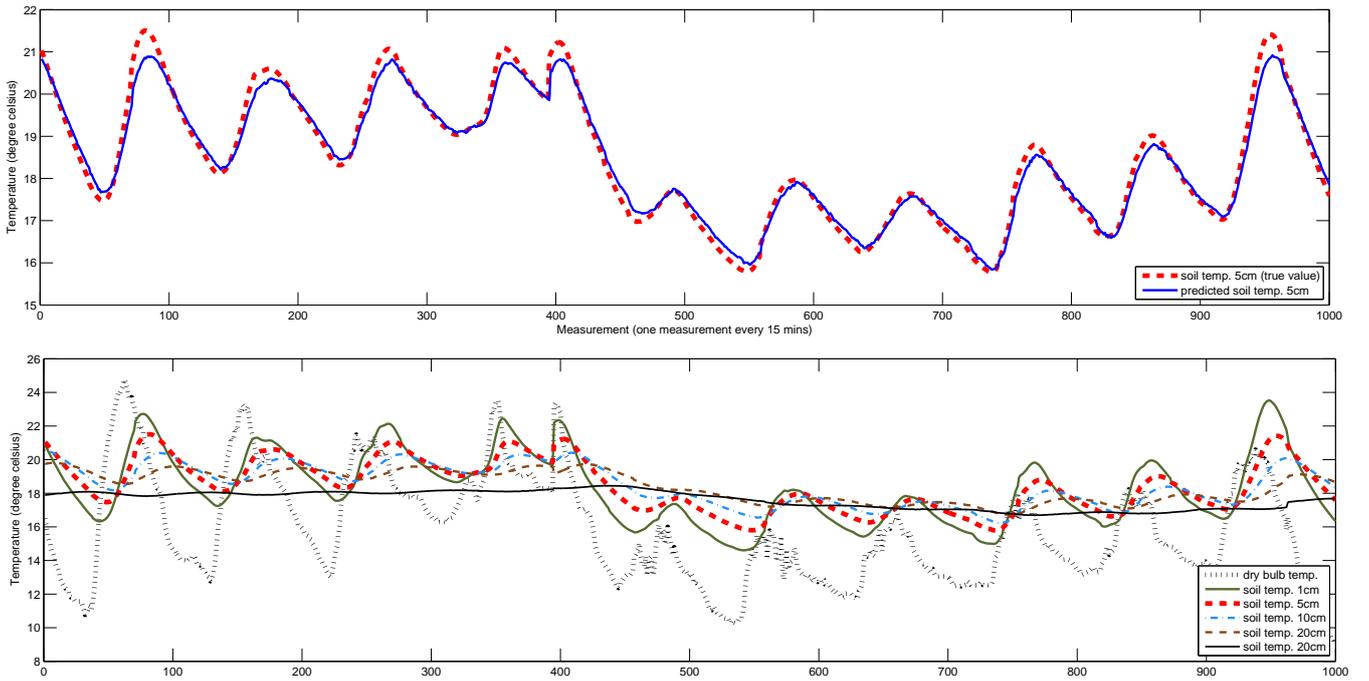} 
\caption{Example of a prediction of the current soil temperature at 5cm (the solid blue line in the top graph, the true value is shown as a dotted red line). The prediction is based on inputs to other sensor nodes (bottom graph). The dotted red line in the bottom graph also shows the soil temperature at 5cm for comparison (not used in the prediction). Additional inputs were radiation and infrared (omitted for clarity).}
\label{fig:prediction3}
\end{figure*}

\subsection{Experimental setup}

\paragraph{Experiment 1 --- amount of training data} 
A number of parameters play a role in training and using SODESN, such as the amount of training data,
the number of units on each node, connectivity between units, link qualities between nodes, etc. In a first
experiment, we used 15 internal units on each node with totally approximately 10\% connectivity between nodes. 
We used a spectral radius of 0.66 for the connectivity matrix, a link quality of 90\% during both training 
and testing, and an increasing amount of training data to obtain learning curves using an incremental 10-fold cross 
validation. The training data varied from 300 data points, corresponding to slightly more than 3 days 
worth of data, up to 30.000 data points, i.e.\ data from a period of 10 month. 
The test data set had a size of 16.665 data points in all cases. 
For each individual experiment, a new SODESN was generated.

\paragraph{Experiment 2 --- reservoir size} 
To evaluate if and how much an increasing number of internal units contributes to higher prediction quality,
we varied the number of internal units per sensor node from 3 to 39 units, resulting in SODESN with 24 up to
312 internal units.
We used a training data size of 30.000 points for training and 16.665 data points for 
testing in a 10-fold cross validation. The basic procedure and all other parameters remained unchanged from 
the first experiment.

\paragraph{Experiment 3 --- ESN vs. SODESN} 
To compare SODESN against a baseline, we simulated a fault detection with global communication
using one (centralized) ESN for each sensor in the network. The ESN we used had 120 internal units each, 
equivalent to a SODESN with 15 units on each of our 8 nodes, and simulated link qualities from 10\% to 100\%
during both training and testing. In the centralized setting, these link qualities represent the quality of the link from 
sensor to central node (independent of the number of hops).

In contrast to using SODESN, in a setup with one ESN for each sensor it is possible to use input data from only 7 
sensors, predicting an 8th sensor of our sensor network. 
\begin{figure*}[hbt]
\centering
\includegraphics[width=\textwidth]{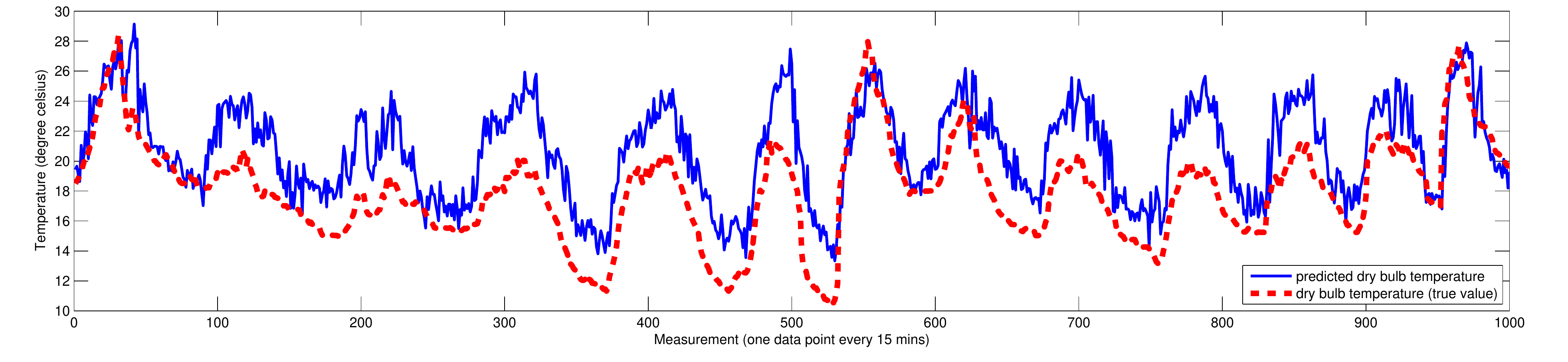} 
\caption{Example of a prediction of the current air temperature as solid blue line, and the true value is shown as a dotted red line. As in Fig.~\ref{fig:prediction3}, the prediction is based on inputs to other sensor nodes.}
\label{fig:prediction1}
\end{figure*}

\paragraph{Experiment 4 --- robustness} 
Sensors in our first experiments deliver time series from different (yet correlated) phenomena, such as 
temperatures at different depths and radiation. To test SODESN with closer correlated inputs,
we computed different time series based on the air temperature data by randomly shifting
the original series up to $\pm 30$ minutes in time and adding uniform random noise of up to $\pm 10\%$. 
(a) A series of tests to determine the prediction quality was run using 8 sensors in a $2 \times 4$ grid.
(b) Then, we extended the size of the sensor network to 100 nodes, arranged in a $10 \times 10$ grid. 
Using these 100 nodes,
we simulated multiple sensor faults to test the effect on the prediction quality for other sensors.
To this end, we randomly selected an increasing number of sensors to fail. Instead of the true value, 
faulty sensors constantly returned zero and fed this value into the SODESN. (c) Finally, we
tested the effect of multiple sensor faults, where faulty sensors were stopped feeding their values into the 
neural network. 
As discussed above, the predictions of their true values as computed by the SODESN were used instead.

\subsection{Results}

\paragraph{Experiment 1}
Figure~\ref{fig:results1} gives an impression of the NRMSE we obtained dependent on the amount of 
training data used. Results are shown for two of the sensors, and 
an overall average NRMSE for all 8 sensors. With an increasing amount of training data,
prediction of our SODESN becomes more reliable, after an initial oscillating phase of 3000 data points.
Table~\ref{tab:exp1} shows NRMSE and some absolute maximum errors of predictions on 
test data. In particular for smaller training sets, absolute errors of the 
more dynamic time series, such as the air temperature, can become quite large for a short period
even though prediction and true value are close over longer intervals.
In this case, the NRMSE between predicted readings and actual values over a window of time
might be a more reliable fault indicator. For less dynamic time series, such as the different soil
temperatures, both NRMSE and absolute errors are small and may be used to indicate faults.

Figure~\ref{fig:prediction3} shows the result of 
a continuous prediction of the soil temperature at 5cm depth, while the sensor for this variable 
fed just random noise into the SODESN during the whole period (slightly more than 10 days). Similarly, the graph in 
Fig.~\ref{fig:prediction1} is a plot of the prediction of the air temperature during the same period, again while
replacing the true temperature measurement by random noise in the input to the SODESN.

\begin{table}[tdp]
\caption{NRMSE and some maximum absolute errors for varying training set sizes.}
\centering
\footnotesize
\begin{tabular}{rr@{.}lr@{.}lr@{.}lr@{.}lr@{.}lr@{.}l}
  & \multicolumn{12}{c} {NRMSE (max. abs. errors in brackets in $^\circ C$)} \\
training & \multicolumn{4}{c}{air} & \multicolumn{4}{c}{soil temp.} & \multicolumn{2}{c}{soil} & \multicolumn{2}{c}{} \\
 set size & \multicolumn{4}{c}{temp.} & \multicolumn{4}{c}{5cm} & \multicolumn{2}{c}{20cm} & \multicolumn{2}{c}{radiat.} \\ \toprule
1500 &     6&76Ê& (155&4) &  0&14 & (1&9)  &   2&64  & 3&40  \\
15000 &     1&11Ê& (26&6) &   0&04 & (1&0) &    0&46  & 0&97  \\
30000 &     0&56Ê& (14&2) &   0&04 & (0&8) &    0&19  & 0&79  \\
\end{tabular}
\label{tab:exp1}
\end{table}%

\begin{table}[bp]
\caption{Some NRMSE and maximum absolute errors for different reservoir sizes from 3 to 39 units per node.}
\centering
\footnotesize
\fcolorbox{cornsilk2}{cornsilk2}{%
\begin{tabularx}{0.47\textwidth}{rcr@{.}lr@{.}lcr@{.}lr@{.}lr@{.}lcr@{.}l} 
 \multicolumn{16}{l} {NRMSE, and max. abs. errors in $^\circ C$ (in brackets)} \\
 & \multicolumn{6}{c}{} & \multicolumn{6}{l}{soil temperature} & \multicolumn{3}{c}{} \\ \cmidrule{8-13}
units & & \multicolumn{4}{c}{air temperature}  & & \multicolumn{4}{l}{5cm} & \multicolumn{2}{l}{20cm} & & \multicolumn{2}{c}{radiation} \\ \midrule
        3 & & 0&67Ê&     (18&2) & &    0&07 &     (1&2) &     0&12 &      &  0&74 \\
      15 & & 0&51Ê&     (15&4) & &    0&04 &     (0&8) &     0&21 &      & 0&76 \\
      27 & & 0&48Ê&     (12&0) & &    0&06 &     (1&2) &     0&15 &      &  0&73 \\
      39 & & 0&47Ê&     (11&6) & &    0&09 &     (2&1) &     0&15 &       & 0&73 \\
\end{tabularx}
 }
\label{tab:exp2}
\end{table}%

\paragraph{Experiment 2}
On average, an increasing number of internal units in the reservoir of our SODESN did not significantly 
improve the prediction quality. Figure~\ref{fig:results2} 
is a plot of the NRMSE for several reservoir sizes from 3 to 39 units per node.
It can be seen from both the plot and from Table~\ref{tab:exp2}
that the prediction of air temperature seems to benefit
from an increased number of units. In other cases, using more internal units does not lead to smaller errors, 
and in some, the error increased even slightly. The average NRMSE over all sensors for SODESN with 312 internal units
is only slightly lower than the average NRMSE for SODESN with only 24 internal units.

\newsavebox{\tempboxiii}%
\begin{figure*}[htp]
\centering
\sbox{\tempboxiii}{\includegraphics[width=0.32\textwidth]{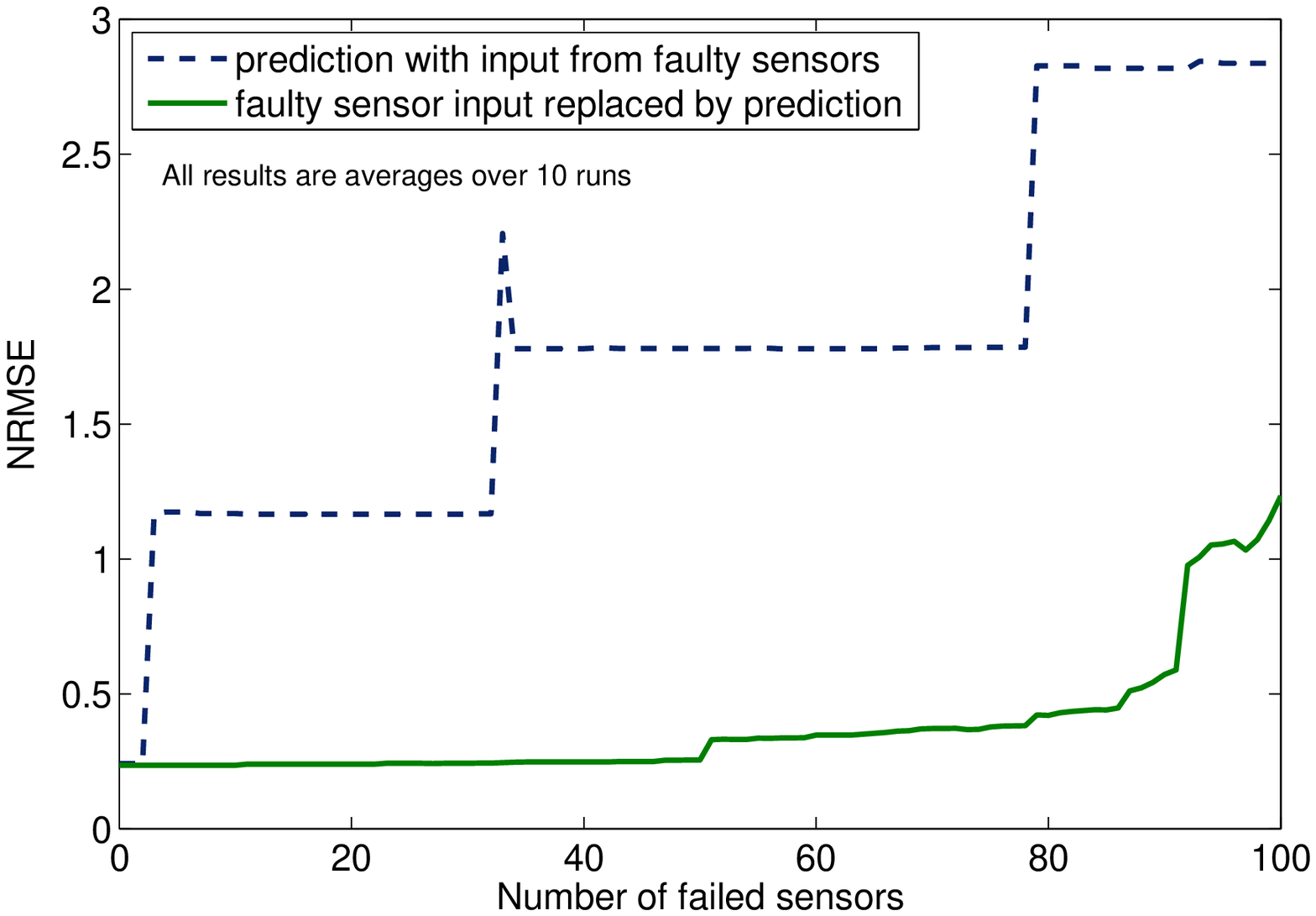}}
\begin{minipage}{0.32\textwidth}%
\subfigure[Performance of the system with increasing number of failures, with and without
replacing faulty sensors with their prediction.]{\usebox{\tempboxiii}\label{fig:exp4-1}}
\end{minipage}%
\hfill%
\begin{minipage}{0.32\textwidth}%
\subfigure[Sample prediction for one sensor with 60 out of 100 sensors failed, and continuing to feed their values into the SODESN.]{%
\vbox to \ht\tempboxiii{%
\vfil
\includegraphics[width=\textwidth]{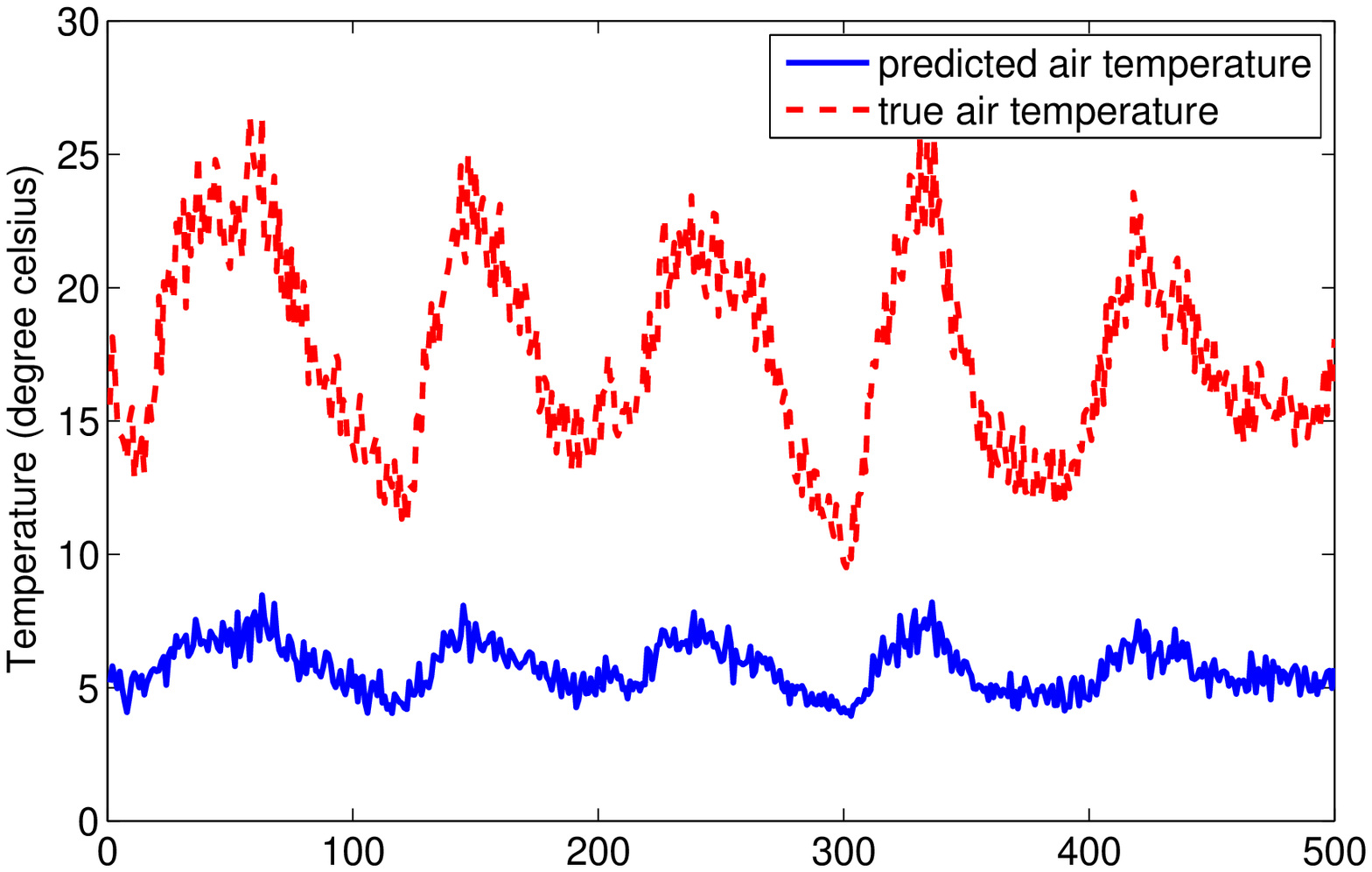}\label{fig:exp4-2}%
\vfil }
}
\end{minipage}%
\hfill%
\begin{minipage}{0.32\textwidth}%
\subfigure[Sample prediction for one sensor, with 60 from 100 sensors failed, and replacing their values by predictions. ]{%
\vbox to \ht\tempboxiii{%
\vfil
\includegraphics[width=\textwidth]{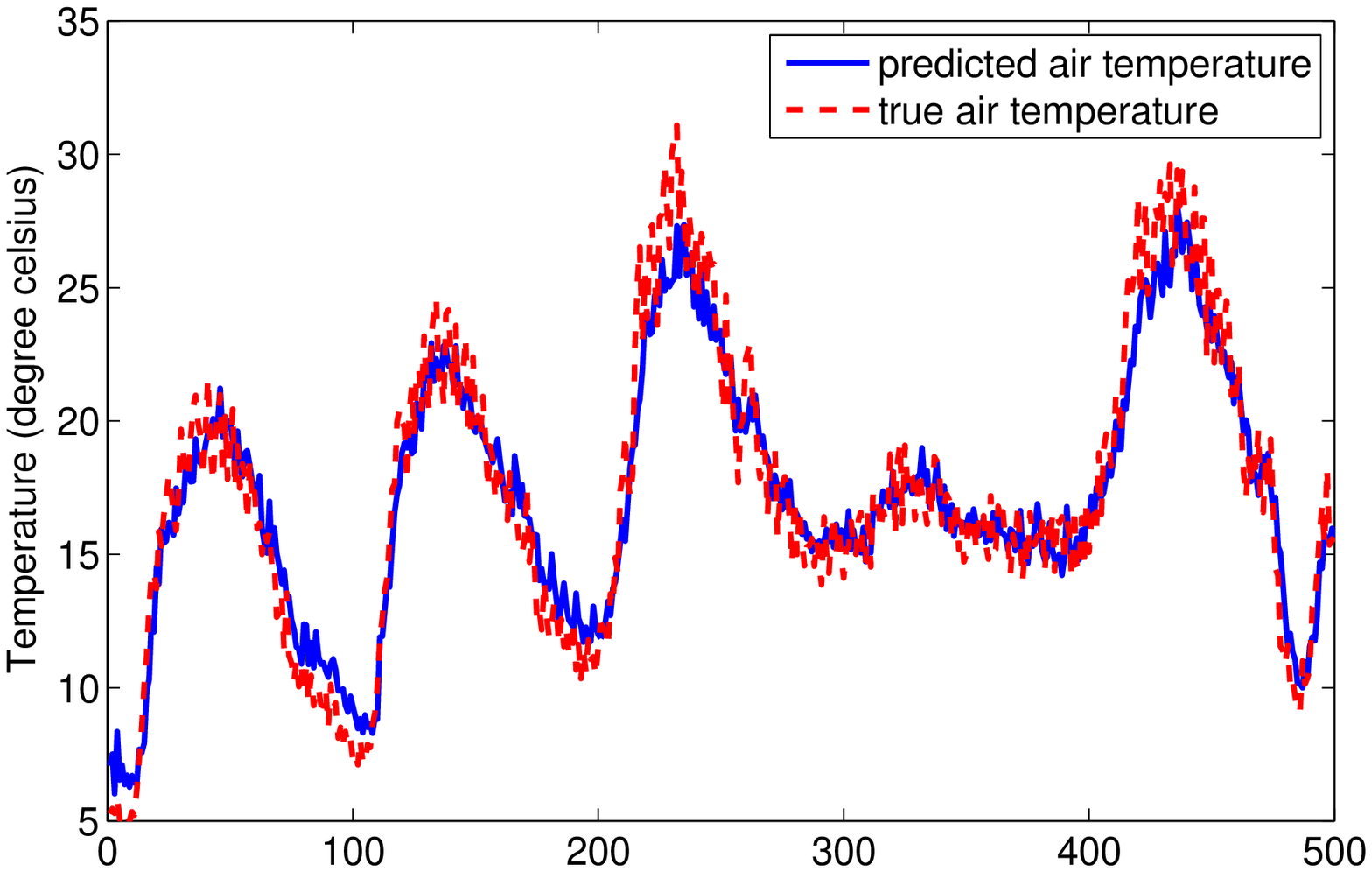}\label{fig:exp4-3}%
\vfil }
}
\end{minipage}%
\caption{Results of various experiments using SODESN and a benchmark using ESN.}
\label{fig:results}
\end{figure*}

\paragraph{Experiment 3}
With a decreasing link quality, the accuracy of the centralized approach using one ESN for each predicted sensor 
decreases rapidly (see Table~\ref{tab:exp3}). 
In contrast to that, SODESN can maintain the same level of accuracy even with poor 
link qualities between local nodes. The graph in Fig.~\ref{fig:results3} shows that the ESN can achieve 
results close to SODESN only under almost perfect conditions. This seems surprising at first,
but the difference in performance is a result of the different methods to pass on sensory information:
In a centralized approach, loss of data has much bigger impact on the result because the missing information
is not replicated elsewhere. In our distributed approach, data is broadcasted to several neighbors (2 or 3 
neighbors in our 8 node experiment, up to 4 neighbors in our experiments with 100 nodes). Because 
in our experiments links between nodes fail independently, the information lost as a result of one link 
failing may still be present in the network and can be used for prediction.

\begin{table}[bp]
\caption{NRMSE in a centralized approach using one ESN with 120 internal units compared to NRMSE 
of SODESN under varying WSN link qualities from 10 to 98\%.}
\centering
\footnotesize
\fcolorbox{cornsilk2}{cornsilk2}{%
\begin{tabularx}{0.47\textwidth}{rcr@{.}lr@{.}lcr@{.}lr@{.}lcr@{.}lr@{.}l}
 \multicolumn{16}{l} {NRMSE in ESN (E) and SODESN (S)} \\
 \multicolumn{7}{c}{} & \multicolumn{9}{l}{soil temperature}   \\ \cmidrule{8-16}
  & & \multicolumn{4}{l}{air temperature}  & & \multicolumn{4}{c}{5cm} & & \multicolumn{4}{c}{20cm} \\
  \cmidrule{3-6}  \cmidrule{8-11} \cmidrule{13-16}
link \% & & \multicolumn{2}{c}{E} & \multicolumn{2}{l}{S} & &
\multicolumn{2}{c}{E} & \multicolumn{2}{c}{S} & &
\multicolumn{2}{c}{E} & \multicolumn{2}{c}{S} \\
\midrule
       10 & &    1&41Ê& 0&51 & & 1&72 & 0&04 &  &   1&83 & 0&19 \\ 
      50 &  &   0&89Ê&  0&54 & & 1&00 & 0&04 &   &  1&04 &  0&23 \\ 
      90 &   &  0&63Ê&  0&51 &  & 0&50 & 0&04 &    & 0&56 &  0&22 \\ 
      98 &   &  0&55Ê&  0&49 &  & 0&27 & 0&04 &    & 0&32 &  0&17 \\ 
\end{tabularx}
}
\label{tab:exp3}
\end{table}%

\paragraph{Experiment 4}
(a) Using 8 more closely correlated air temperature time series, we achieved an almost constant NRMSE of 0.2 for 
SODESN independent of the number of units (from 3 units/node up to 39 units/node), and a maximum absolute
prediction difference of $6^\circ$C. The lowest NRMSE
in experiment 2,  where we used soil temperatures and radiation data to predict air temperature,
was 0.47 (Table~\ref{tab:exp2}). The better performance in this scenario was expected.
(b) Scaling the experiment up to 100 sensor nodes, the prediction has about the same quality as with only 8 
sensors. Then, we begin to subsequently fail random sensors. A first qualitative (visual) inspection of the 
predicted time series vs. the true values shows acceptable performance up to more than 60\% of failed sensors
(see Fig.~\ref{fig:exp4-2} for a sample prediction with 60 failed sensors). More
quantitatively, from the graph in Fig.~\ref{fig:exp4-1} we see that failing up to 16 of the sensors does not change the
performance of the system at all. In our experiments, the average maximum absolute error for up to 16 failed nodes 
was below $11^\circ$C, and for up to 32 failed nodes, it remained below $16^\circ$C. For 60 failed nodes, the 
NRMSE has grown from 0.26 to about 1.0, with an maximum absolute error of around $19^\circ$C. (c)
Feeding back the predictions of the true value instead of faulty sensor values results in a greatly improved prediction 
quality, so that the average error lower is almost constant for up to 50\% of failed nodes. Even for more than 50\% 
failed nodes, the error increases only slowly until around 90\%.

\section{Discussion}
\label{sec:discussion}

Our first experiment showed that the amount of training data used strongly influences the prediction quality. 
Further aspects seem to be the ``correlatedness'' of different sensors and the dynamics of the time series. 
Some of the ``easier'' sensors in our experiment could be successfully modeled after training on 1500 data points 
($\approx 15$ days of training data), while for ``harder'', less correlated sensors we needed at least 5000 points 
($\approx 52$ days).
Our offline learning approach requires to perform a computation on the whole training time series.
In particular for larger data sets this will usually be done on a machine outside the network.
The learning then computes sets of output weights for each sensor node. A way to deal with less correlated sensors
may therefore be to successively improve the SODESN by re-training on increasingly larger data sets and 
exchanging the learned weights over time.

A second important factor is the amount of local communication introduced. From the description
of our architecture in Sect.~\ref{sec:desn} it follows that neighbors exchange activations of their local internal units
 -- one value per unit and sample step.
Results from our second experiment are therefore interesting, because we have seen that the number of internal 
units did not play a crucial role -- we used only 3 units in some experiments. SODESN communication
and sample rate of sensors does not have to run synchronously with each other. Alternatively it is also possible collect 
some data locally, and to run the SODESN on larger blocks of data, as long as all nodes run their part of the
SODESN at the same rate (proxy units would have to be changed to queues in this case).

The amount of local computation required is similarly dependent on the number of units. In contrast to
the offline training, exploitation requires only a few operations, for each internal
unit a number of additions, multiplications, and computation of $\tanh(x)$.

\section{Conclusions}
\label{sec:conclusions}

In this paper, we presented SODESN, a novel distributed recurrent neural network architecture for 
creating models of dynamical systems. We introduced an offline learning approach for SODESN
that is closely related to training ESN and inherits the low computational complexity of the original
approach. We then presented an approach to train SODESN for fault detection in WSN, where
predictions of sensor values are made based on information from neighbor nodes. 

Our evaluations on real-world data show that our approach can be used to build models of
dynamic time series and help to detect sensor faults. We have shown that the approach is robust to
WSN link failures through its distributed computation and local communication. SODESN outperform 
a comparable, centralized approach assuming realistic link qualities.
Using only local communication
also contributes to SODESN scaling well with an increasing number of WSN nodes. 

We have also shown that our approach is robust against multiple node failures. In our evaluation using the 
predictions of failed sensors as input, 
50\% of the sensors failed without affecting prediction quality, and the performance degraded gracefully 
up to slightly more than 80\% failed nodes.


\bibliographystyle{apalike2} 
\bibliography{literature}

\end{document}